\documentclass[dvipsnames,format=sigconf,anonymous=false,review=false]{acmart}
\usepackage{graphicx}

\usepackage{amsmath,amssymb,amsfonts}
\usepackage{algorithmic}
\usepackage{textcomp}
\usepackage{xcolor}

\usepackage[ruled]{algorithm2e}

\SetAlFnt{\small}
\SetAlCapFnt{\small}
\SetAlCapNameFnt{\small}
\SetAlCapHSkip{0pt}

\begin{document}

\title{Data-Informed Model Complexity Metric for Optimizing Symbolic Regression Models}

\author{Nathan Haut}
\author{Zenas Huang}
\author{Adam Alessio}
\begin{abstract}
  Choosing models from a well-fitted evolved population that generalizes beyond training data is difficult. We introduce a pragmatic method to estimate model complexity using Hessian rank for post-processing selection. Complexity is approximated by averaging the model output Hessian rank across a few points (N=3), offering efficient and accurate rank estimates. This method aligns model selection with input data complexity, calculated using intrinsic dimensionality (ID) estimators. Using the StackGP system, we develop symbolic regression models for the Penn Machine Learning Benchmark and employ twelve scikit-dimension library methods to estimate ID, aligning model expressiveness with dataset ID. Our data-informed complexity metric finds the ideal complexity window, balancing model expressiveness and accuracy, enhancing generalizability without bias common in methods reliant on user-defined parameters, such as parsimony pressure in weight selection.
\end{abstract}




\maketitle

\section{Introduction}
Symbolic regression is a powerful technique for discovering mathematical models that describe data, offering advantages over traditional regression methods due to its ability to produce interpretable models. However, symbolic regression faces challenges related to the complexity of the search space and computational efficiency, particularly in high-dimensional settings \cite{tran2016genetic, parsimony}. Genetic Programming (GP), a popular method for symbolic regression, employs evolutionary strategies to navigate the vast search space of potential solutions. There has been significant research in the space of genetic programming to improve the ability for evolution to efficiently find generalizable models. These methods include Lexicase selection \cite{lexicase}, Pareto tournaments \cite{tournament}, using correlation as opposed to RMSE \cite{Haut2023}, using feature construction techniques \cite{evofeat}, among many others.

\begin{figure}[h]
\centering
\includegraphics[width=8cm]{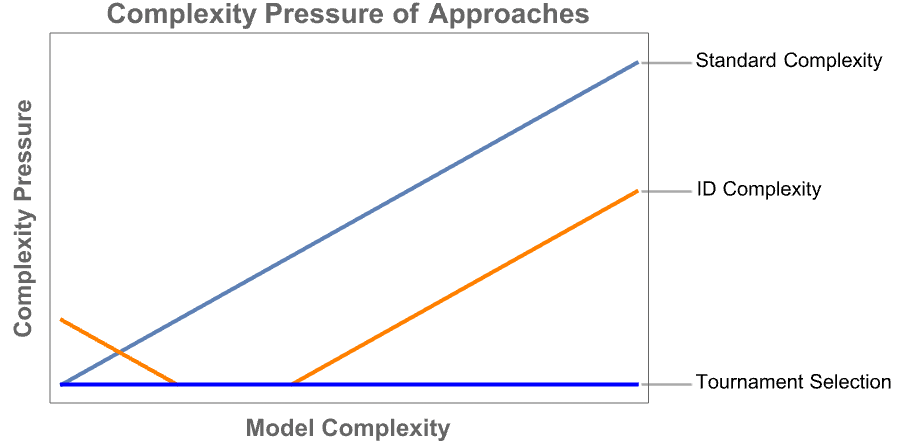}
\caption{{\bf Complexity Pressure Comparison.} Shown here are the complexity pressures imposed by tournament selection and standard complexity metrics used in Pareto tournament selection and parsimony pressure compared to our proposed complexity metric. The plateau in our metric represents the target ID range since our approach does not assume a single point of optimal complexity.}
\label{fig:pressure}
\end{figure}

While using the standard Pareto tournament selection approach where selection promotes models that have a balance between complexity and accuracy \cite{tournament}, one of the potential limitations is the implicit assumption of zero complexity to be optimal. Unfortunately, this results in populations that preserve many models with low complexity and also low fitness, which occupies population space without offering useful genetic information. We hypothesize that information about the ID of the data will identify a more appropriate range of non-trivial complexities to help select models that are neither too simple nor too complex. To highlight this, Figure \ref{fig:pressure} shows the complexity pressure of our proposed approach compared to the complexity pressure imposed by both standard tournament selection and traditional complexity metrics such as in Pareto tournament selection.

In this work, we explore the inclusion of ID estimate ranges as a post-processing step in the genetic programming framework to enable us to identify an optimal subset of Pareto front models that have a model expressivity that is aligned with the data's native complexity as estimated by ID. This work is intended as a first step towards understanding how aligning model complexity with data ID impacts generalizability. Our contributions are summarized as: 
\begin{enumerate}
\item We implement a new model complexity metric for aligning model complexity with data dimensionality.
\item We demonstrate a sampling strategy to efficiently compute our method. 
\item We describe a post-processing procedure for selecting models from an evolved population using our metric.
\item We demonstrate that our method can identify models which are more likely to generalize well. 
\end{enumerate}

\section{Background}
Genetic Programming (GP) has been widely used in symbolic regression for its ability to evolve complex mathematical expressions that fit data. Traditional GP approaches often produce excessively complex models when dealing with high-dimensional data due to a lack of guidance toward simpler solutions. This has led to the development of various approaches to promote simplicity such as parsimony pressure \cite{parsimony} and Pareto tournament selection \cite{tournament}. These approaches adapt selection pressure to reward models that are simpler. While these approaches have helped address issues of bloat and overfitting, they unfortunately introduce the risk of underfitting since they tend to significantly reward models with near-zero complexity. Furthermore, since excessively simple models are preserved in the populations, these simple low-fitness models can occupy significant portions of the population space and thereby slow evolutionary progress. 

ID refers to the minimum number of variables or degrees of freedom required to effectively describe the underlying structure of a dataset. In other words, ID aims to capture the true complexity of the underlying low-dimensional manifold that describes the data by identifying the key latent factors contributing to the data's variation \cite{camastra_intrinsic_2016}. For high-dimensional datasets, the ID is typically much lower than the observed dimensionality, offering a more realistic measure of the data's inherent complexity. ID estimation offers a promising solution to this challenge by capturing the true complexity of data by identifying the degrees of freedom that most contribute to variation in the data. By integrating ID estimates into the genetic programming search process, we aim to prioritize models aligned with the data's intrinsic structure, potentially reducing unnecessary complexity and enhancing generalization.

We employ various ID estimation algorithms from the Scikit-Dimension library, which offers a broad view of a data's intrinsic complexity through diverse estimators based on common geometric, statistical, and topological properties of data \cite{Bac_2021}. By applying each of these estimators to the datasets for symbolic regression problems from the Penn Machine Learning Benchmark (PMLB), we aim to evaluate how ID impacts model discovery and the performance of discovered models \cite{romano2021pmlb}.

\section{Methods}
\begin{algorithm}
\caption{model selection by ID-ED alignment}
\SetAlgoLined
\KwData{Dataset $D$}
\KwResult{High quality model subset}

Compute ID distribution profile using scikit-dimension:\;
$ID_{mean}, ID_{stdev}, \gets \text{ComputeIDProfile}(D)$\;
$ID_{max} \gets ID_{mean} + ID_{stdev}$\;
$ID_{min} \gets ID_{mean} - ID_{stdev}$\;
$\text{models} \gets \text{FitModelsWithStackGP}(D)$\;

\For{model $m$ in models}{
    $p_1, p_2, p_3 \gets \text{GetStrategicPoints}(m)$\;
    $H_1 \gets \text{ComputeHessian}(m, p_1)$\;
    $H_2 \gets \text{ComputeHessian}(m, p_2)$\;
    $H_3 \gets \text{ComputeHessian}(m, p_3)$\;
    $\bar{H}_m \gets \frac{H_1 + H_2 + H_3}{3}$\;
    $ED_m \gets \text{Rank}(\bar{H}_m)$\;
}

$\text{highQualityModels} \gets \{m : ID_{min} \leq ED_m \leq ID_{max}\}$\;
\Return{highQualityModels}\;
\end{algorithm}

\subsection{Datasets}
For our experiments, we used the symbolic regression datasets from the Penn Machine Learning Benchmark (PMLB). These datasets represent a broad range of intrinsic complexities, making them suitable for evaluating how ID estimates can inform the symbolic regression process. 

\subsection{Intrinsic Dimensionality Estimation}
To estimate the ID of each dataset, we utilized multiple algorithms from the scikit-dimension Python library \cite{Bac_2021}. 
Below we provide a brief description of each of the ID estimators available from scikit-dimension which were used to establish our target ID ranges for each symbolic regression dataset from the Penn Machine Learning Benchmark.
\begin{itemize}
    \item Correlation Dimension (CorrInt): Analyzes how the number of data points scale with a varying neighborhood radius by calculating the correlation integral $C(r)$ which is a count of pairs of points within distance r of each other. The ID is then estimated based on how $C(r)$ scales with $r$ in the limit. \cite{GRASSBERGER1983189, RJ-2017-054}.
    \item Dimensionality from Angle and Norm Concentration (DANCo): Utilizes the concentration of normalized distances between points and the concentration of angles formed by triplets of data points and uses max likelihood estimation to find an ID estimate that best explains both properties \cite{ceruti2012dancodimensionalityanglenorm}.
    \item Expected Simplex Skewness (ESS): Examines the geometry of simplices formed by nearby points and estimates the dimension by analyzing the distribution of simplex skewness i.e. how "stretched" these simplices are, which follow patterns based on the ID \cite{6866171}.
    \item Fisher Separability Algorithm (FS): Estimates the ID by analyzing how well points can be separated by random hyperplanes. The separability changes systematically with the manifold dimension due to concentration of measure phenomena in high dimensions \cite{albergante2019estimatingeffectivedimensionlarge}. 
    \item K-Nearest Neighbors Algorithm (KNN):  Provides an ID estimate based on minimum spanning trees and analysis of the distribution of distances to k-nearest neighbors. This method relies on how the ratios of distances to consecutive neighbors follow patterns that depend on ID \cite{5233815}.
    \item Principal Component Analysis (lPCA): Analyzes how many principal components are necessary to explain the variance in local neighborhoods around each point and uses that value as the ID estimate. Depending on the user hyperparameter settings, different variants of lPCA exist for choosing the number of significant components. \cite{cangelosi_component_2007, fan2010intrinsicdimensionestimationdata, 1671801}.
    \item Manifold-Adaptive Dimension Estimation (MADA): Estimates dimension locally around data points using nearest neighbor techniques and combines these local estimates. The main idea is that the convergence rate of these estimates depends only on the ID of the manifold and not the ambient space dimension \cite{inproceedings}.
    \item Minimum Neighbor Distance Maximum Likelihood Algorithm (MiND ML): Uses maximum likelihood estimation on the ratios of distances between points and their nearest neighbors. These ratios follow specific distributions which are contingent on the ID \cite{Rozza2012NovelHI}.
    \item Maximum Likelihood Estimation (MLE): Analyzes how the number of data points within a small radius scales as that radius varies. Uses maximum likelihood estimation to find the dimension that best explains this scaling behavior \cite{Levina2004MaximumLE, Haro2008TranslatedPM, Hill1975ASG}.
    \item Method of Moments (MoM): Uses extreme value theory to analyze the distribution of minimum distances between points which follow patterns based on the ID \cite{Amsaleg2018ExtremevaluetheoreticEO}.
    \item Tight Local Intrinsic Dimensionality Estimator (TLE): Uses extreme value theory in order to provide tight local estimates of dimension. Relies on analyzing how neighborhood sizes vary across the dataset to estimate the local dimension \cite{amsaleg2022intrinsicdimensionalityestimationtight}.
    \item TwoNN Algorithm: Relies on analyzing just the ratio of distances to first and second nearest neighbors for each point. This ratio follows a specific distribution which depends on the ID \cite{Facco_2017}.
\end{itemize}

For each dataset, these algorithms produced a range of twelve ID estimates due to the differing underlying geometric assumptions about the data manifold presupposed by each algorithm. Nevertheless, these ranges allowed us to compute a mean and standard deviation of IDs which enabled us to define a plausible range for the true ID of the dataset and thus served to provide guidance for what we believed the appropriate level of model complexity should be as measured by our novel Hessian rank-based metric.

\subsection{Genetic Programming Search with StackGP}
In this work, we utilize the StackGP system as the symbolic regression engine for developing models \cite{StackGPRepo}. StackGP is a stack-based genetic programming implementation in Python that represents regression models using two stacks, one for the model's operators and one for the model's input features and constants. 

The parameter settings used with StackGP are shown in Table \ref{tab:parameters}. We chose the same values as were used in \cite{10700803}. During evolution, $R^2$, Pearson's correlation coefficient was used as the fitness metric, and combined stack length was used as the complexity metric. We used $R^2$ as the fitness metric since it was previously found to make evolution more efficient when compared to RMSE \cite{Haut2023}. 

\begin{table}
\caption{StackGP Parameter Settings\label{tab:parameters}}
\begin{center}
\begin{tabular}{ll} 
\hline
Parameter & Setting\\
\hline
 Mutation Rate & 79 \\ 
 Crossover Rate & 11 \\
 Spawn Rate & 10\\
 Elitism Rate & 10\\
 Crossover Method \hspace{5mm}  & 2 Pt.\\
 Tournament Size & 30\\
 Population Size & 400\\
 Generations & 200 \\
 Max Complexity & 300 \\
 Fitness Metric & $R^2$ \\
\hline
\end{tabular}
\end{center}
\end{table}

\subsection{Measuring Model Complexity}

Our approach consists of two complementary components working in tandem: (1) ID estimation methods that determine an ideal target complexity range for a given dataset, and (2) a Hessian rank-based metric that measures the actual complexity of evolved models. While in Section 3.2, we described how we used various ID estimation methods to establish target complexity ranges for datasets, this section introduces our novel Hessian rank-based complexity measure. The ID estimates serve to guide what model complexity to target ("how complex should our models be"?), while our Hessian rank metric assesses the achieved complexity ("how  complex is this particular model?"). This focus on functional rather than structural complexity aligns with prior work \cite{10.1007/978-3-642-20407-4_3, Le2016ComplexityMI, 10.1145/1830483.1830643, VANNESCHI2021114929, 10.1007/978-3-642-20407-4_23, 8790341, cravens2024geneticprogrammingexplainablemanifold} in genetic programming and symbolic regression that emphasizes measuring how models behave rather than how they are constructed.


Drawing on related work using the Fisher Information Matrix of a model to measure geometric complexity \cite{novak_geometric_2021, abbas2021effectivedimensionmachinelearning}, we elected to examine the rank of a model's Hessian matrix as a computationally efficient measure of a model's effective dimensionality (ED), i.e. the number of significant directions of variation that the model expresses. Thus, to quantify the complexity of each of our evolved models we implemented a numerical approximation of the average Hessian of the model, evaluated at three points in the input space corresponding to the mean, minimum, and maximum target values. The rank of this Average Hessian was then computed to provide an estimate of each model’s capacity to represent the complexity intrinsic to various datasets as reflected by the pre-computed ID estimates. By comparing the rank of the average Hessian with the range of IDs estimated for each dataset, we were able to identify an optimal set of models within each evolved Pareto front whose EDs were well aligned with the data's own intrinsic complexity as estimated by ID. 

The Hessian of a model is a matrix of the second-order partial derivatives of the model's output with respect to its parameters, providing insight into how the model's predictions respond to small changes in the parameter values within a local neighborhood. A high rank in the Hessian indicates that the model is sensitive in many independent directions in the parameter space, implying a more complex model at that point. Conversely, a lower rank suggests that the model's predictions depend on fewer parameters, indicating a simpler model with fewer directions of variation at a given point. While the Hessian is itself a measure of local curvature, by evaluating the Hessian at three points in the input space: mean, minimum, and maximum of target values, and averaging the Hessian, we are able to obtain a more representative summary of the model's curvature across the domain (as available in the training data) instead of being limited to function behavior at a single point. The rank of the averaged Hessian then gives us a rough approximation to what we expect to be the model's ED which we can compare to the ID estimates to assess how well the model's complexity is aligned with the data's intrinsic structure.

It is important to note that our method measures complexity by counting the number of independent directions of variation in a model, not by measuring the magnitude of curvature in any particular direction. For example, a function like $sin(x)$ would be identified as having complexity 1 despite its high curvature because it varies in only one independent direction. Conversely, a linear function like $x_{1} + x_{2} + x_{3}$ would be identified as having complexity 0 by our metric, because despite varying in three first-order directions, it has no second-order variation (all second order derivatives are zero). In this latter case, evolution would be unnecessary since a linear regression would easily find the solution. In either of these examples, our approach would still select the appropriate corresponding model complexity: in the former, we would select for single-variable functions while in the latter case, our ranged approach would tend towards choosing either constant or linear models. Thus, the rank of the Hessian captures this geometric notion of independent directions of second-order variation, making it an appropriate metric for measuring the ED of a model. 

\subsection{Theoretical Framework for Model-Data Dimensionality Alignment}
\subsubsection{Model Effective Dimensionality and Hypothesis}
We define a model's effective dimensionality (ED) as the number of independent variables needed to capture all significant variations in the model's input-output mapping. More formally, given a model $f:\mathbb{R}^n\mapsto\mathbb{R}$, its effective dimensionality $d$ is the smallest number such that there exists a function $g: \mathbb{R}^{d}\mapsto\mathbb{R}$ and a mapping $\varphi:\mathbb{R}^{n}\mapsto\mathbb{R}^{d}$ where $\|f - g \circ \varphi\|<\epsilon$ is an approximation holding for some sufficiently small error tolerance $\epsilon > 0$. This notion reflects the true functional complexity of a model, independent of its parametric form so that a model which appears to depend on many variables but whose output effectively varies only along a few key directions would have a low effective dimensionality.

Our core conjecture is that a model's ED should align with the ID of the dataset it describes. More precisely, we hypothesize that, for a well-fitted model $h$ with training error below some threshold $\tau > 0$, its ED should approximately equal the data ID.

\subsubsection{Theoretical Requirements}
For this ID-ED relationship to be valid, the following conditions are necessary:
\begin{enumerate}
    \item Manifold structure: there exists some $d$-dimensional Riemannian manifold $\mathcal{M}\subset\mathbb{R}^n$ and a function $f:\mathcal{M}\mapsto\mathbb{R}$ such that the data points $(x,y)$ satisfy $y = f(x) + \varepsilon$, where $\varepsilon > 0$ represents observation noise and the manifold $\mathcal{M}$ must have constant dimension $d$ almost everywhere.
    \item Manifold parametrization: there must exist a finite atlas of charts $\{(U_{i}, \varphi_{i})\}$ where each $\varphi_{i}:U_{i}\mapsto\mathbb{R}^{d}$ is a $C^{k}$ diffeomorphism with $k\geq 2$ and the transition maps between charts are sufficiently smooth to ensure that the manifold structure may be learned from the data.
    \item Sampling Adequacy: the data sampling distribution $P(x)$ must have support over $\mathcal{M}$ with sufficient density to resolve $d$-dimensional features. That is, $\forall \epsilon > 0 \hspace{0.1cm} \exists \delta>0$ such that any ball of radius $\delta$ in $\mathcal{M}$ has enough sample points to estimate the local dimensionality within $\epsilon$.
\end{enumerate}
Under conditions 1-3, for any model $h$ with training error below $\tau$ that approximates $f$ within $\epsilon$ (i.e.$\text{sup}\lVert h(x) - f(x)\rVert <\epsilon$ for $x\in\mathcal{M}$), then there exists a constant $C > 0$ such that $\lVert \text{rank}(H_{h(x)}) - d \rVert < C\epsilon$ for all $x\in \mathcal{M}$ where $H_{h(x)}$ is the Hessian of $h$ at $x$.

\subsubsection{Mathematical Justification}
To establish why the rank of the Hessian matrix provides an appropriate model complexity metric, we present the following mathematical framework. Consider a fitted model $f:\mathbb{R}^{n}\mapsto \mathbb{R}$ to be a twice-differentiable function that varies primarily along a $d$-dimensional smooth manifold $\mathcal{M}\subset \mathbb{R}^{n}$, where $d<n$. At each point $x\in \mathcal{M}$, the Hessian matrix is given by:
\[ H(x) := \left[ \frac{\partial^2 f}{\partial x_i \partial x_j} \right]_{i,j=1}^n\]
and at each point $x\in\mathcal{M}$, there exists a $d$-dimensional tangent space $T_x\mathcal{M}$. Locally, we can decompose the Hessian relative to the tangent and normal directions of the manifold so that:
\[ H(x) = D^2 f|_{T_x\mathcal{M}} + D^2 f|_{N_x\mathcal{M}}\]
where $D^2 f|_{T_x\mathcal{M}}$ is the restriction of the Hessian to the tangent space and $D^2 f|_{N_x\mathcal{M}}$ is its orthogonal complement. If $f$ is well-fit to the training data, then it captures the manifold structure and will vary primarily along $\mathcal{M}$ and the contributions from $D^2 f|_{N_x\mathcal{M}}$ will be neglibible or zero. Thus, at any point $x\in\mathcal{M}$ the rank of $H(x)$ will be determined by $D^2 f|_{T_x\mathcal{M}}$ and we will have $\text{rank}(H(x))\leq d$. If $f$ exhibits meaningful curvature in all directions of $T_x\mathcal{M}$, then $\text{rank}(H(x)) = d$.
\subsubsection{Asymptotic Analysis and Practical Considerations}\label{theory}
Next, consider the average Hessian matrix $\bar{H}_{N}$ computed over $N$ points $\{x^{i}\}\subset\mathcal{M}$:
\[ \bar{H}_N = \frac{1}{N} \sum_{i=1}^N H(x^{i})\] 
Then for finite $N$, $\text{rank}(\bar{H}_{N})$ may be less than $d$ due to incomplete coverage of the manifold's curvature directions. However, as $N$ increases, the sampling becomes more comprehensive and $\text{rank}(\bar{H}_{N})$ increases accordingly. Under suitable conditions of smooth tangent spaces and non-degenerate second derivatives, the eigenvalues of the Hessian will maintain consistent signs when projected onto the tangent space such that we may expect rank to be preserved during averaging thereby ensuring the averaged Hessian will accumulate contributions from all independent curvature directions implying  
\[ \lim_{N \to \infty} \text{rank}(\bar{H}_N) = d  \]

In practice, we observed that the actual datasets from the Penn benchmark often exhibited varying local dimensionality across different scales - a violation of the constant dimension assumption. Rather than invalidating our approach, we account for this by:
\begin{enumerate}
    \item Using multiple ID estimation methods that capture dimensionality at different scales
    \item Defining a target range around the mean ID estimate
    \item Employing strategic sampling of points (minimum, maximum, and mean target values) to efficiently approximate global model complexity
\end{enumerate}

In addition, although the theoretical framework assumes differentiability, our practical implementation uses numerical approximations to handle the broader class of functions typically evolved in GP. We implement finite difference methods to approximate second derivatives, allowing us to handle non-differentiable functions and discontinuities which may appear in evolved expressions. Importantly, the accuracy of the computed Hessian matrix itself is less critical than obtaining a reliable estimate of its rank as our metric aims to count the number of independent directions of variation rather than obtain a precise measurement of curvature in any particular direction. This aligns with our goal of measuring functional complexity by assessing the number of independent ways a model can vary, rather than the magnitude of those variations. This approach provides a practical framework for aligning model complexity with data structure while acknowledging the challenges of real-world datasets.

\section{Results}

\subsection{Analysis of ID Estimators}

\begin{figure}[h]
\centering
\hspace*{-0cm}\includegraphics[width=8.5cm]{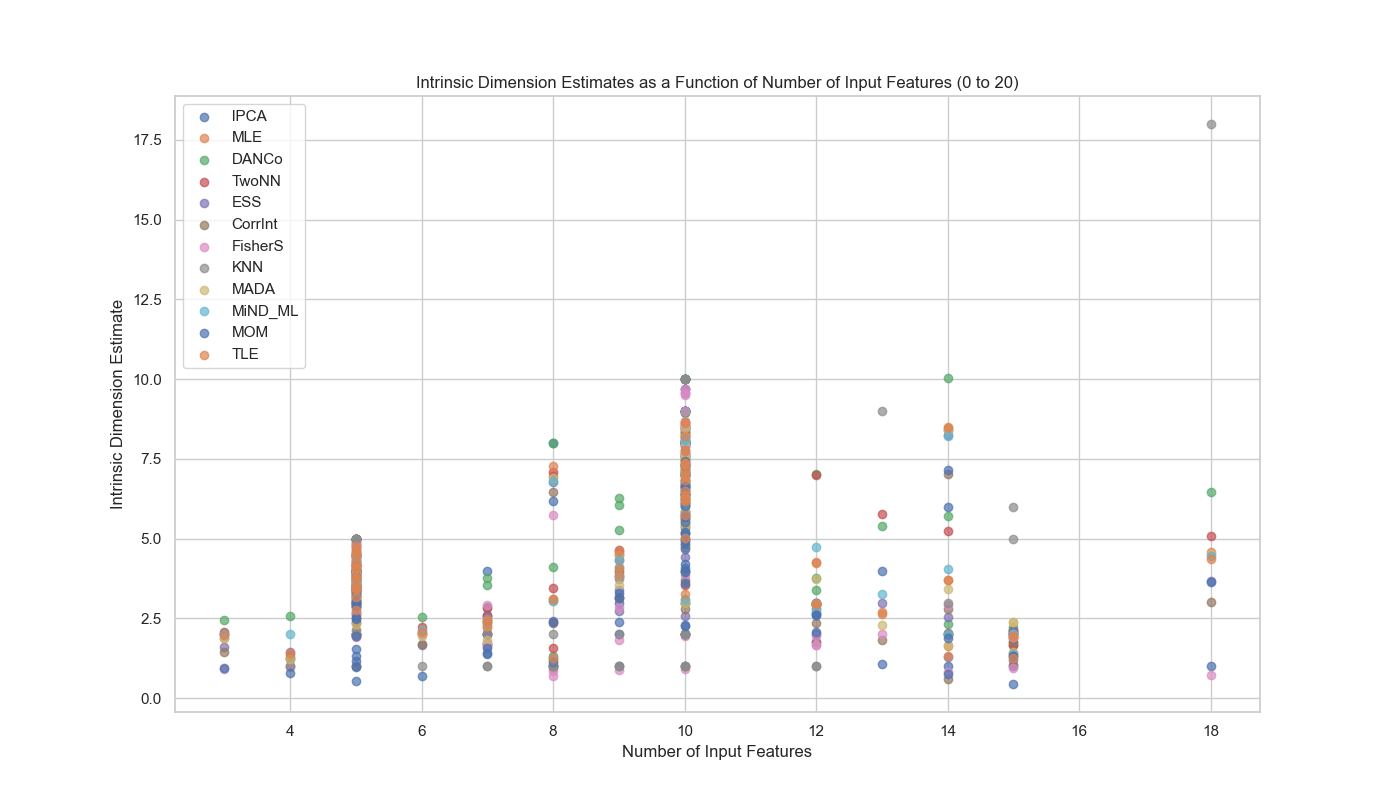}
\caption{{ID estimates by algorithm across datasets of different number of input features demonstrating the high variance in estimated ID on datasets with similarly sized input feature counts.}}
\label{fig:IDbyFeatCounts}
\end{figure}

\begin{figure}[h]
\centering
\includegraphics[width=8cm]{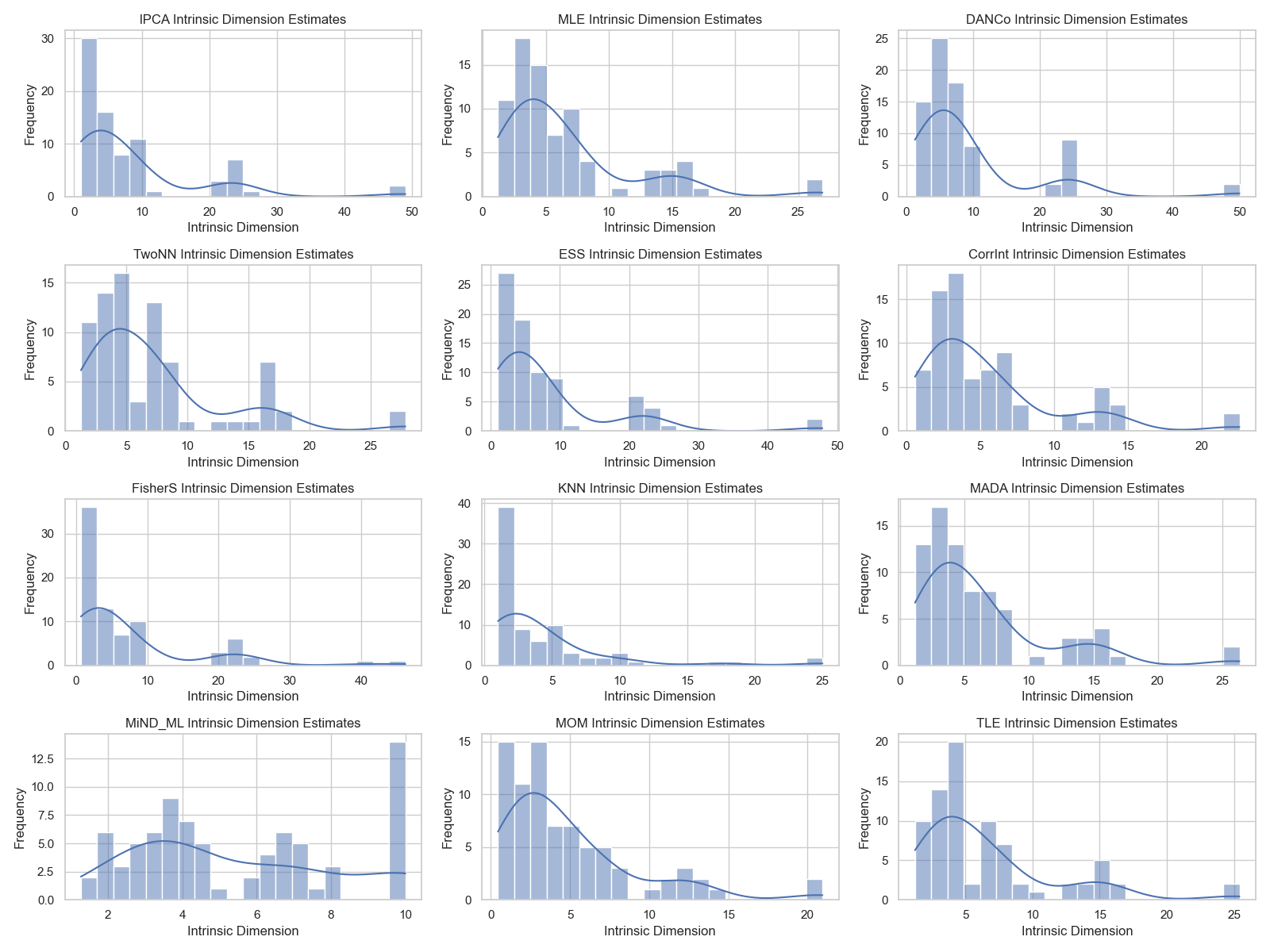}
\caption{Distribution of ID estimates for each method on all datasets further highlighting the variance across methods.}
\label{fig:IDbyMethod}
\end{figure}

Analysis of the empirically computed ID estimates revealed substantial variability across both algorithms and datasets (See Figure \ref{fig:IDbyFeatCounts} and \ref{fig:IDbyMethod}). For example, while KNN and MADA produced conservative estimates (typically below 10), lPCA and DANCo often suggested much higher dimensionality for the same datasets which reflected the different geometric assumptions of each method and suggested the presence of varying local IDs. ID distributions by method were generally right-skewed but varied significantly in scale and spread. This variability challenged the determination of a single "true" ID.

Notably, the relationship between a dataset's feature count and its estimated ID is not straightforward. Datasets with similar numbers of features often yielded markedly different ID estimates, indicating that raw feature count alone is an inadequate proxy of a dataset's true complexity. For example, among datasets with 50-100 features, we typically observed ID estimates ranging from 10 to 50, suggesting that, in line with the manifold hypothesis, many of these datasets could be described by lower-dimensional manifolds despite their high number of input features, thus motivating our proposed approach to pre-characterizing dataset complexity with ID estimation.

Given the uncertainty in ID estimates as can be seen in Figure \ref{fig:IDbyFeatCounts}, rather than a single ID value we elected to instead utilize a range of ID estimates for each dataset (i.e. $\pm{1}$ standard deviation of the mean ID value computed across all 12 estimators) as an ideal window for identifying models of optimal complexity. By considering the distribution of estimates, we sought to establish plausible windows of model complexity that were neither too restrictive nor too permissive with respect to the task of capturing the data manifold's true complexity. 

The observed variability in ID estimates across different methods may also be understood through the lens of manifold multiplicity. While the manifold hypothesis suggests that high-dimensional data lies on or near a lower-dimensional manifold, this manifold is not necessarily unique. Recent work has also suggested that the standard manifold hypothesis is incomplete and that a collection of distinct manifolds describing the same data at different scales and complexities may be a more empirically accurate characterization of many high-dimensional datasets \cite{brown2022the, brown2023verifyingunionmanifoldshypothesis}. The potential for a multiplicity of manifolds has two important implications for our work. First, it helps explain why different ID estimation methods may produce varying results for the same dataset since each method probes the dataset's structure at different scales and complexities utilizing different assumptions resulting in the detection of distinct manifold structures with different dimensionalities (e.g. The lPCA estimator assumes certain linear relationships in the data are significant while the MLE method assumes data distributed within neighborhoods according to a Poisson process) \cite{cangelosi_component_2007, Levina2004MaximumLE}. Second, manifold multiplicity may also explain the observation that the GP algorithm can evolve a variety of models with distinct estimated complexities while still fitting the data well. Just as different ID estimation methods may detect different manifold structures, the evolved models of varying complexity may reflect different successful strategies for representing the data, each aligning with a different manifold structure that describes the variation in the data.

Thus, given the potential for multiple manifolds, the variability in ID estimates observed can be understood as not merely a consequence of estimator disagreement arising from noise or sampling error but a reflection of the genuine complexity of the underlying geometric structures in the data. This understanding also reinforces our decision to use a range of ID estimates rather than seeking a unique "true" ID, as the use of an ID range is more flexible in allowing for the selection of a plurality of models capable of identifying the multiplicity of intrinsic manifold structures which are equally effective as parsimonious descriptions of the data whereas the choice of a single best ID value would likely be too restrictive.

While the Hessian calculation traditionally requires differentiability, as mentioned in \ref{theory}, we implement numerical approximations to handle the broader class of functions typically evolved in GP. Our implementation uses a finite difference methods to approximate second derivatives, allowing us to handle non-differentiable functions and discontinuities which may appear in evolved expressions. While this is an approximation of the theoretical framework presented in section 3.5, our empirical results demonstrate that it provides a reliable measure of model complexity in practice and can be reasonably well-aligned with the estimated ID ranges (accounting for the sampling error and any potential miscounting of the Hessian rank which may arise from underestimating the number of directions due to selection of non-differentiable points or insufficient sampling).

\subsection{Current Selection Methods}
\begin{figure}[h]
\centering
\hspace*{-0.0cm}\includegraphics[width=8.5cm]{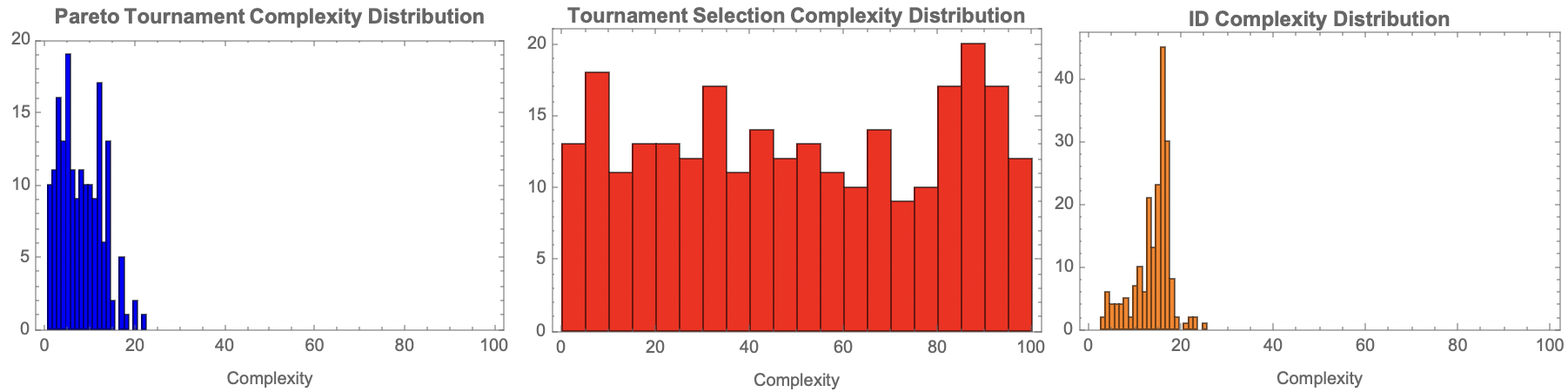}
\caption{Complexity distributions of a model population when using Pareto tournament selection, standard tournament selection (complexity \& accuracy), and when incorporating ID as a third objective for Pareto tournament selection using the "auto\_price" dataset.}
\label{fig:dists}
\end{figure}

\begin{figure}[h]
\centering
\includegraphics[width=8cm]{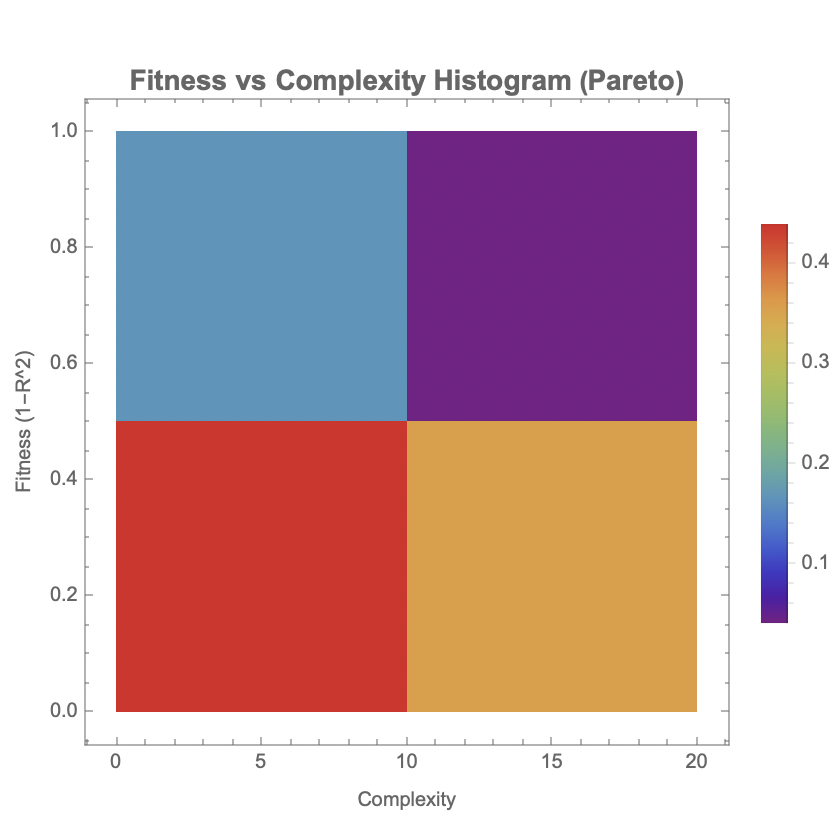}
\caption{{\bf Complexity vs Fitness (1-$R^2$) Histogram for Pareto Tournament Selection.} Shown here is a histogram representing the distribution of models in a population after evolution which used Pareto tournament selection. The results show that a significant portion fall into the upper left quadrant which represents models which are overly simple and not very accurate. }
\label{fig:dists2}
\end{figure}

To explore how our metric would influence the complexity distributions when used for selection rather than just post-processing selection we conducted a single trial incorporating our metric into an evolutionary run (Note, our metric is currently too computationally expensive to feasibly run a benchmark on all problems). Figure \ref{fig:dists} shows the complexity distributions of populations when using Pareto tournament selection and standard tournament selection compared to our approach using the data-informed complexity metric to select models. In this case, our approach was used in a modified Pareto tournament selection where size complexity was replaced with our model dimensionality metric. The results show that Pareto tournament selection has a heavy bias towards zero complexity which results in a significant portion of the population space being occupied by models that are excessively simple (underfit). This is further demonstrated in Figure \ref{fig:dists2} which shows that about 20\% of models selected by Pareto tournaments selection fall into the very simple, low-fitness region, thus occupying population space that could be better utilized. Tournament selection does not have any regard for complexity, so a uniform distribution is observed which introduces overfitting risk since there is no penalty for increasing complexity to achieve marginal improvements in training performance. Further, when using standard tournament selection, there is risk of losing interpretability of models by evolving models with unnecessary complexity. The distribution of our data-informed metric is controlled and prioritizes models of a specific complexity range. It is interesting to note, that while the data-informed complexity metric is not directly related to the standard size-based complexity metric which is the x-axis in these plots, there does appear to be a clear pattern when observing the distribution with respect to standard size-based complexity since the x-axis in each plot is size complexity. 

When applying our method, we first use ID estimators to establish target complexity ranges for each dataset. These ranges inform the selection of models based on their Hessian-derived complexity measures. For example, if ID estimators suggest a dataset has ID between 3-5, we preferentially select models whose Hessian rank-based complexity measures fall within this range. This two-step process helps ensure that model complexity aligns with the underlying complexity of the data being modeled. 

\subsection{Data Informed Complexity}
\begin{figure}[h]
\centering
\includegraphics[width=6cm]{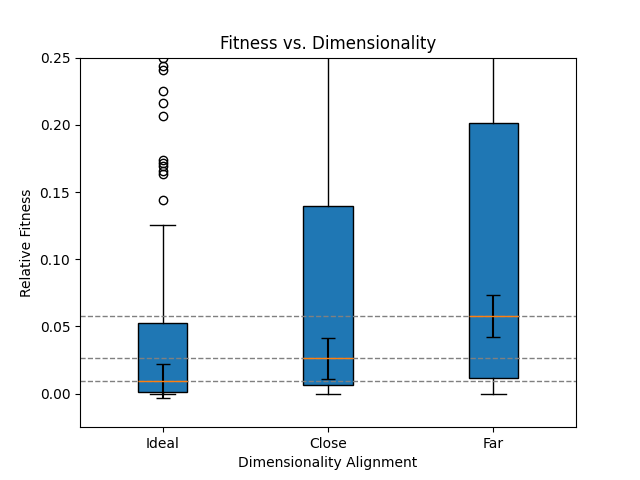}
\caption{Normalized performances of evolved Pareto front models optimized for all 121 data sets with respect to their measured data-informed complexity where "Ideal" models are selected within the target model dimensionality range using the data-informed metric, "Close" complexity contains models within $\pm{1}$ dimensionality measure using our metric, and "Far" are outside of the targeted dimensionality range by greater than 1 unit. The results show that the models within the targeted complexity range generally have better (lower) test fitnesses. In each bar, the median fitness is indicated by the orange line, with standard errors reported around the median by the black bars. The models in the "ideal"
 range reported a median normalized fitness of $0.010\pm{0.013}$, the models in the "close" range reported a median normalized fitness of $0.0213\pm{0.0153}$, the models in the "far" range reported a median normalized fitness of $0.0573\pm{0.0155}$. 
 }
\label{fig:summaryResults}
\end{figure}

To determine whether there is value in using the data-informed complexity metric as a post-processing step to filter models from an evolved population of quality models, we evolved models for all 121 benchmark problems and selected the Pareto front of the models with respect to fitness and standard size complexity. From those models, we then grouped them by their data-informed model dimensionality measure and recorded their relative fitnesses. The results are shown in Figure \ref{fig:summaryResults}. The "Ideal" range indicates models whose complexity was within the target dimensionality range computed from our data ID estimates. The "Close" range indicates models within 1 dimensionality unit from the target range. The "Far" range contains the rest of the models. The results show that the models with the best median performance fall into the "Ideal" range with the models farthest from the target dimensionality range typically performing the worst. The three groups were compared pairwise using the Mann-Whitney statistical test with a significance level of 0.05. The resulting p-values were: ideal vs. close (0.00656), ideal vs. far ($1.62*10^{-7}$), and close vs. far (0.0302). All values are below the significance threshold, indicating that the differences in median values are statistically significant. 

\begin{figure}[h]
\centering
\includegraphics[width=6cm]{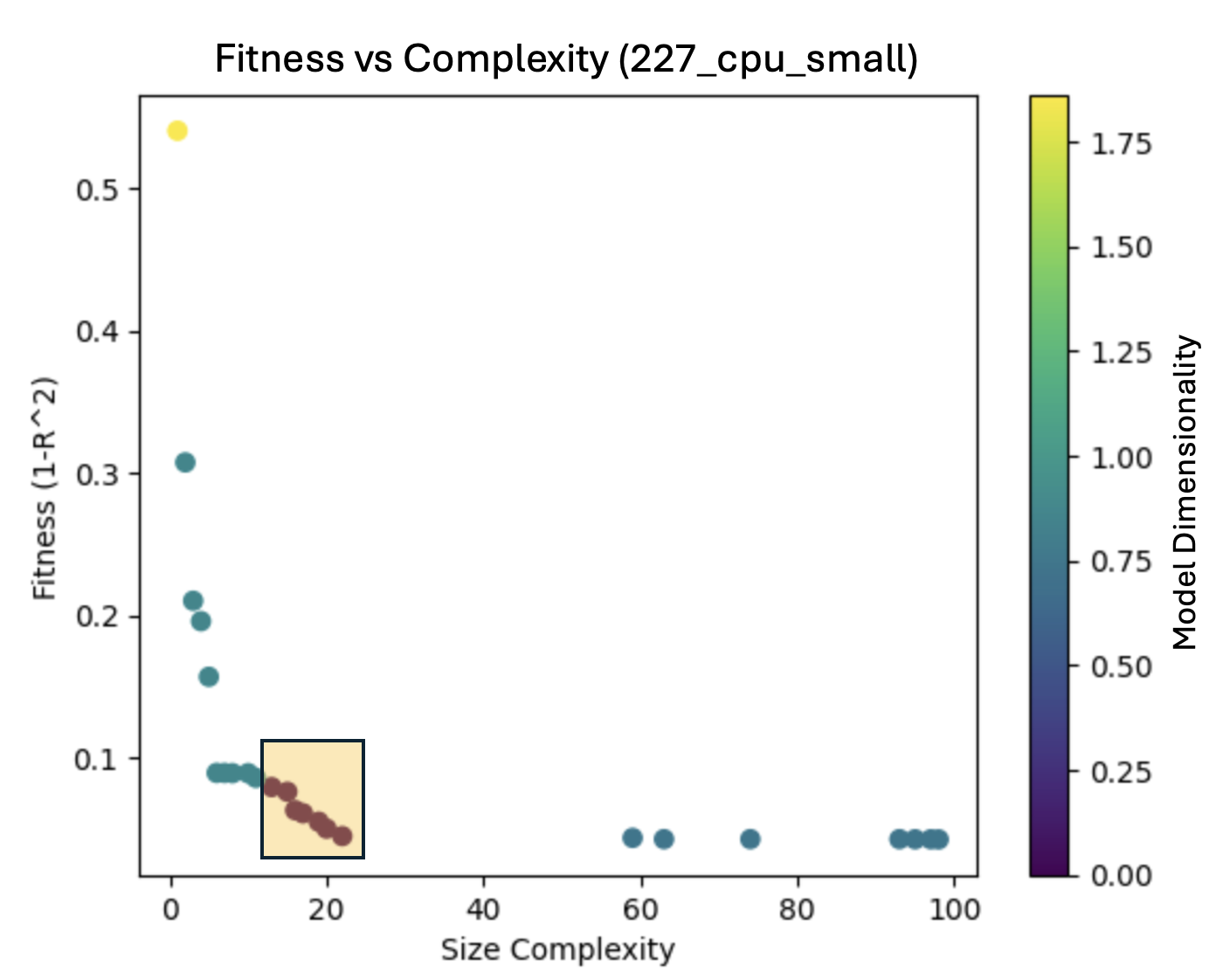}
\caption{Pareto front for evolved models fit to the 227\_cpu\_small dataset. The color indicates the model dimensionality. The yellow box highlights that models selected with the data-informed approach are near the knee of the Pareto front (those with enough complexity to capture the data, but not excess complexity). }
\label{fig:227_front}
\end{figure}

To better visualize which models are selected for inclusion in the "Ideal" range, models within the target dimensionality range, we show the Pareto front of evolved models on the "227\_cpu\_small" dataset in Figure \ref{fig:227_front}. The results show that the models which fall into the "Ideal" range have a good trade-off between accuracy and size complexity with some having comparable performance to the best-performing models in the front yet with substantially less complexity.

These results show that our Hessian rank-based complexity metric can be used as a post-processing step to filter for models in or near the target dimensionality range from the set of high-quality models returned at the end of evolution. By filtering for models in or near the targeted dimensionality range, we can focus on models that are more likely to generalize well without introducing superfluous complexity that risks reducing interpretability. In addition, it reduces the risk of introducing user bias when selecting which model to use from the evolved Pareto front by providing a heuristic for narrowing down the candidate pool. 


\subsection{Data Informed Complexity Metric Sensitivity}
\begin{figure}[h!]%
    \centering
    [\centering  A]{{\includegraphics[width=5cm]{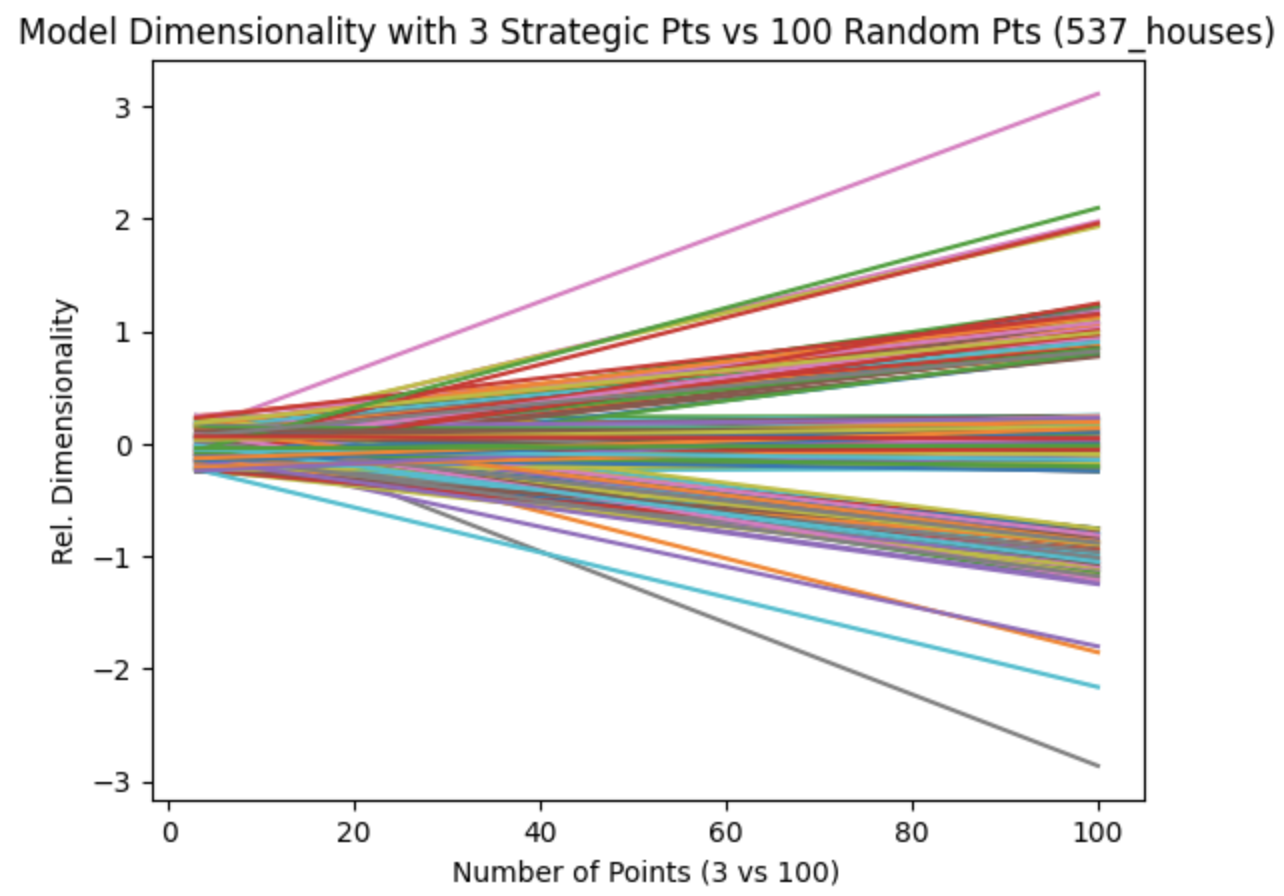} }}%
    \qquad
    [\centering B]{{\includegraphics[width=5cm]{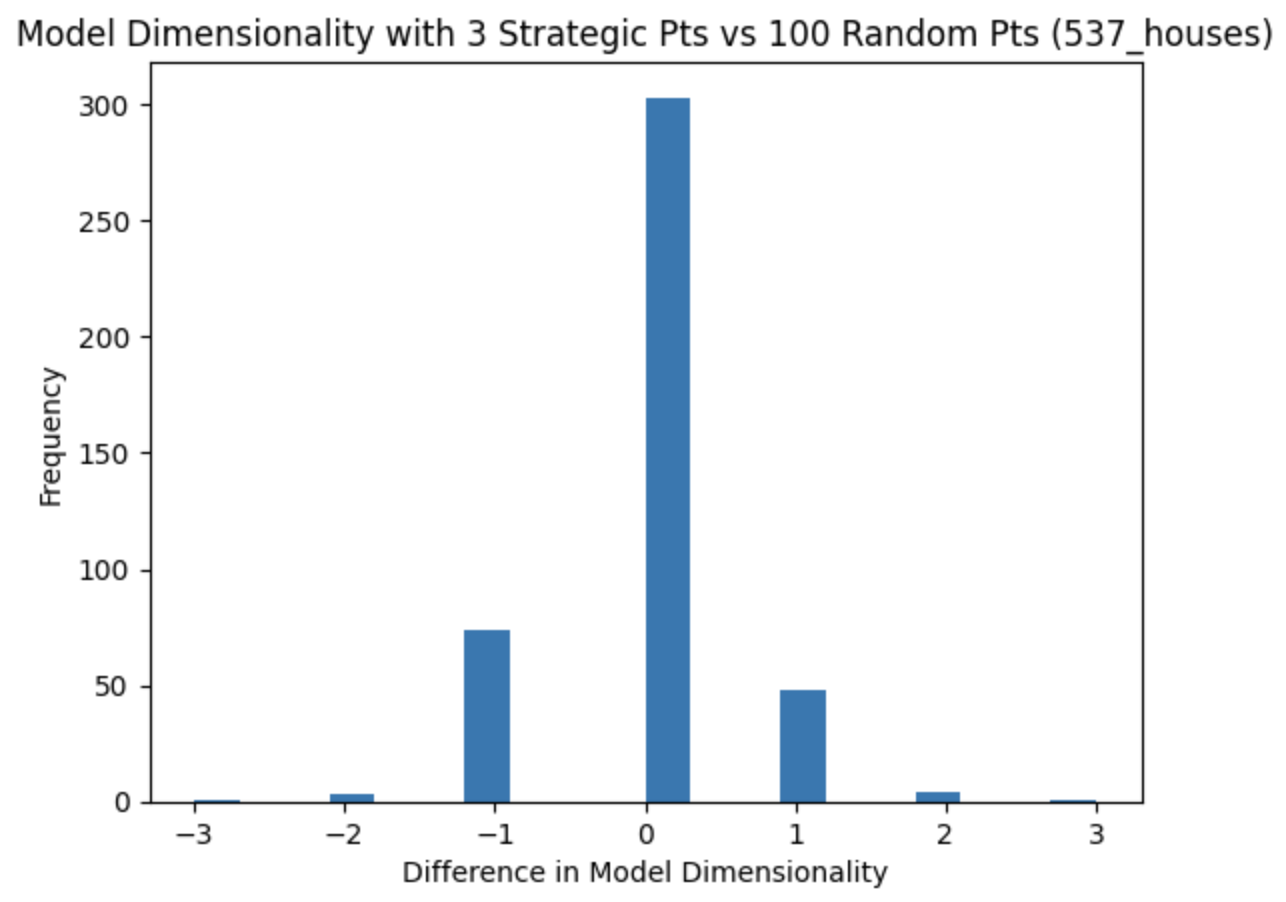} }}%
    \qquad
    \caption{{\bf[A] Three Strategic Pts vs 100 Random Pts} The results show that frequently the three strategically selected points find the same model dimensionality as 100 randomly selected points. {\bf [B] Difference in Model Dimensionality (3 Strategic Pts vs 100 Random Pts)} The results show that in about 70\% of cases, the 3 points we select strategically leads to the same dimensionality computed when using 100 random points across the model surface. Further, in about 98\% of cases we are within $\pm{1}$ for the computed dimensionality.}%
    \label{fig:tuningPerf}%
\end{figure}

In our experiments, we utilized three strategically selected points to sample the model surface (Hessian calculation) to approximate the model dimensionality. These three points were chosen with the goal of sampling a diverse range of the data while using few points to keep the dimensionality approximation relatively fast. The points were chosen as the max, min, and mean points with respect to the target (response) value of the data to obtain adequate coverage of model behavior across the input domain. To determine if our choice of three strategically sampled points was reasonable, we compared the computed model EDs using our three strategic points as well as model EDs from using 100 randomly sampled points for the 537\_houses dataset. 
Figures \ref{fig:tuningPerf} A and B show the comparison between estimated dimensionalities of our three strategically sampled points compared to 100 randomly sampled points across a model population of around 400 models. The results show that our approach with three points achieves the same estimated dimensionality as 100 randomly sampled points in about 70\% of cases and is within $\pm{1}$ dimension in about 98\% of cases, indicating that our strategic sampling is reasonably effective at a fraction of the computational cost.

\section{Discussion of advantages of Hessian-rank complexity versus Traditional Complexity Metrics}

While our Hessian rank-based complexity metric requires additional computation compared to simpler metrics like model size, it provides several key advantages that justify its use:

\begin{enumerate}
    \item Semantic Understanding: Unlike syntactic measures that count nodes or operations, our metric captures \textit{functional} complexity of the models. For example, two models of identical size may have very different functional behaviors -- our metric can distinguish between them while size-based metrics cannot.
    \item Alignment with Data Complexity: As demonstrated in section 4.3 and Figure \ref{fig:summaryResults}, models selected using our metric which align with data ID estimates show improved generalization compared to models selected using purely size-based complexity measures. Models in the "Ideal" category (those with an ED most aligned with the ID profile of the dataset) achieved a median normalized fitness of 0.010 $\pm$ 0.013, significantly outperforming models outside the "Ideal" range, indicating that our metric is useful for filtering quality models from an evolved population.
    \item Computational cost: while computing the Hessian rank is more expensive than explicit model size, we show in Section 4.4 that our three strategically sampled points provides a reliable approximation. This makes the computational overhead manageable, especially since in this work we are primarily interested in this metrics' application as a post-processing step rather than during evolution. 
    \item Mitigation of User Bias. By measuring functional complexity instead of model size we provide a principled way of model selection based on the complexity that is inherent to the data, removing the risk of bias introduced by user expectations of model complexity that are often based on subjective assessments of appropriate complexity.   
    
\end{enumerate}

These results show that while our method requires additional computation, the improved model selection and generalization performance justify this cost for applications where model quality is critical. Future work could explore ways to further optimize the computation of this metric while maintaining its benefits and potentially its integration as a selection pressure in the evolutionary process.

\section{Conclusion}
In this work, we introduced a model complexity metric based on counting the number of significant directions of variation expressed by a model as given by the rank of a numerically approximated average Hessian matrix. This metric enables methods for selecting models that have a complexity that matches more closely with the ID of the input data. We also showcased the risks of using standard complexity pressures during model evolution since they may promote the selection of models that have near zero complexity, resulting in underfitting and reducing evolutionary efficiency by occupying population space with overly simple models. We then explored a range of data dimensionality metrics for capturing the ID of the data and found that these methods can show significant variance on the same data. This observation led us to avoid reliance upon any single estimate of data dimensionality and instead opt for using a range of acceptable target model dimensionalities informed by our ID estimates. We then demonstrated how our model dimensionality metric can be used to select models in the computed target range, and further, that those models which are in or near the range tend to perform better than models far from it. This indicates that our metric has utility as a post-processing step to select which models from an evolved population should be deployed. Finally, we also demonstrated that by using three strategically sampled points on the model surface we can efficiently approximate our model dimensionality metric.

In future work, we plan to explore how our model dimensionality metric can be further computationally optimized and used to influence evolution to guide GP search towards development of models with sufficient yet not excess model complexity. Since our metric is still slower to compute relative to standard complexity, introducing it as a direct replacement is not reasonable given typical performance expectations, which is why we did not include that exploration in this work. We also plan to incorporate additional means of accounting for local ID differences in order to address the manifold multiplicity and scale-related ID variation that we observed in our data's ID estimates, so as to further refine the target complexity ranges we obtained in this work.


\subsubsection{Code \& Data Availability}
Code and data will be made available via a Github link after anonymous submission stage. 

\bibliographystyle{ACM-Reference-Format}
\bibliography{references}

\end{document}